# EigenCoin: sassanid coins classification based on Bhattacharyya distance

Rahele Allahverdi [a] *, Mohammad Mahdi Dehshibi [a], Azam Bastanfard [b], Daryoosh Akbarzadeh [c]

[a] *Department of Electrical, Computer and IT, Islamic Azad University, Qazvin Branch, Iran*
[b] *Department of Computer Engineering, Islamic Azad University, Karaj Branch, Iran*
[c] *National Museum of Iran, Tehran, Iran*

**Abstract**

Solving pattern recognition problems using imbalanced databases is a hot topic, which entices researchers to bring it into focus. Therefore, we consider this problem in the application of Sassanid coins classification. Our focus is not only on proposing EigenCoin manifold with Bhattacharyya distance for the classification task, but also on testing the influence of the holistic and feature-based approaches. EigenCoin consists of three main steps namely manifold construction, mapping test data, and classification. Conducted experiments show EigenCoin outperformed other observed algorithms and achieved the accuracy from 9.45% up to 21.75%, while it has the capability of handling the over-fitting problem.

## 1. Introduction

In recent years, much research has been devoted to make the pattern recognition algorithms applicable and solve the real world problems. Many desired applications such as biometric data analysis, medical image processing, quantifying industrial products, and optical/handwritten character recognition are highly demanded by this requirement. However, cultural heritage studies have been less taken into consideration due to several constraints including imbalanced distribution of data and limited number of the suitable data in which the illumination, occlusion and intra-class variations observe carefully.

[*] ADDRESS FOR CORRESPONDENCE: Rahele Allahverdi. *Department of Electrical, Computer and IT, Islamic Azad University, Qazvin Branch, Iran . E-Mail:* mohammad.dehshibi@ieee.org

Ancient coins classification is a branch of cultural heritage study, which its automation has advantages of reducing the processing time and increasing the processing speed and accuracy. Moreover, as the great majority of the stolen coins are sold through the Internet and manual tracking of this trade is fairly impossible, utilizing pattern recognition methods for analyzing such coins is a necessity.

As it is mentioned above, problems related to the data set are causal factors for decreasing the number of researches in this area. In [1], different local descriptors are applied to the coins images by Kampel et al. and it was observed in the course of experiments that SIFT descriptor has the capability of producing a promising result on the coins classification. Arandjelovic [2] also investigates the SIFT descriptor for recognizing Roman's coins; however, the end result would seem to indicate that this method is inaccurate and its recognition rate is about 2.4%. With a view to the visual words and utilizing locally biased directional histogram based visual codebooks, Arandjelovic can achieve the recognition rate of 57% in Roman's coin classification. Although the latter work conducted experiments on a database with 65 classes and examined different features for comparing the end results, it is similar to the Kampel work in performing local features and the experiments are not reproducible. The reason is that the detail of feature's parameters and data distributions corresponded to between/within classes are left in doubt. In addition to ancient coins, several works have been reported on classifying modern coins. For instance, Huber et al. [3] utilize Eigen analysis in order to construct a discriminative model which has the capability of classifying coins from their diameter and thickness. Another identification system for matching EURO coins is presented by Khashman et al. [4] in which a neural network is trained with images relating to the both side of EURO in different rotated positions. In [5], angular and distance information of coin's edge image are encoded in histograms; then, a 3-nearest neighbor approach is utilized on two sides of the coin to construct a classification pilot so-called COIN-O-MATIC.

In this paper, we investigate the problem of Sassanid coins classification with complicated patterns' structure. With the aim of solving this problem, first, a set of training images is utilized to construct a discriminative model by means of Principal Component Analysis (PCA). PCA [16] is a dimensional reduction method, which extracts the most informative principal components of the multi-dimensional data and considers them as basis vectors for constructing a subspace, hereinafter referred to as the EigenCoin. Afterwards a test image is projected onto the EigenCoin and Bhattacharyya distances between its coefficients vector and those representing each training coin are computed. Then the test coin will associate with the minimum distance class. Finally, in order to compute the classification rate and get closer to the actual results, a new metric is defined which has the capability of overcoming the imbalanced distribution of images in the data set. With the aim of demonstrating the efficiency of the proposed method, two sets of experiments are conducted on Sassanid coin images. In the first set, different numbers of Eigenvectors with Bhattacharyya distance are examined in order to achieve the maximum recognition rate. In the second set, we compare proposed method together with the accuracy of three feature extraction and classification methods including Bi-Directional PCA (BDPCA) [6], Wavelet decomposition [7], and Harris corner detector [8]. These results would seem to indicate that the proposed method has the capability of being a pilot for ancient coin classification with chaotic patterns.

The roadmap of this paper is as follows. In Section 2, the proposed method is described. Experimental results are presented in Section 3 and finally, the paper is concluded in Section 4.

## 2. Proposed Method

In the first step, a coin is isolated from its background. Then 70% of data is utilized to construct EigenCoin space. Afterward a new images is mapped onto EigenCoin manifold and its distance with all manifold's member is calculated. Finally, if the result is less than a threshold, the image will be assigned to the class with the minimum distance. The overall process of the proposed method is illustrated in Fig 1.

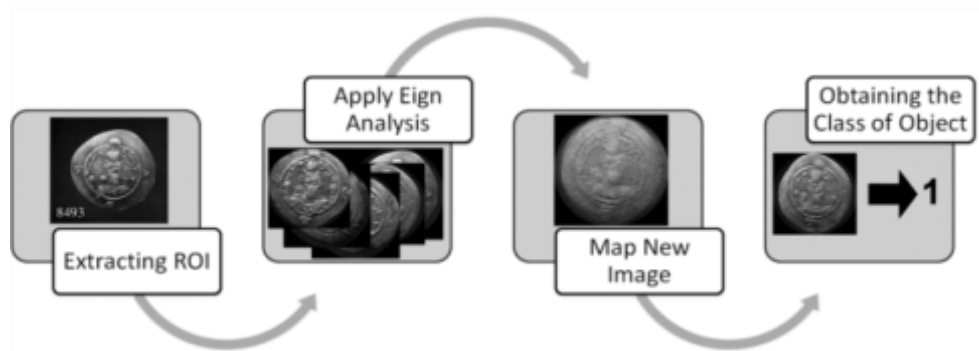

Fig. 1. Procedure of the proposed method

*2.1. Extracting Region of Interest*

The contrast of the coin to be segmented differs greatly from the background, and gradient operators have the capability of detecting contrast changes in the image. Therefore, edge information from Sobel operator is utilized to obtain a binary mask which facilitates the segmentation process. The binary gradient mask highlights lines of high contrast in the image, although these lines do not quite delineate the outline of the coin.

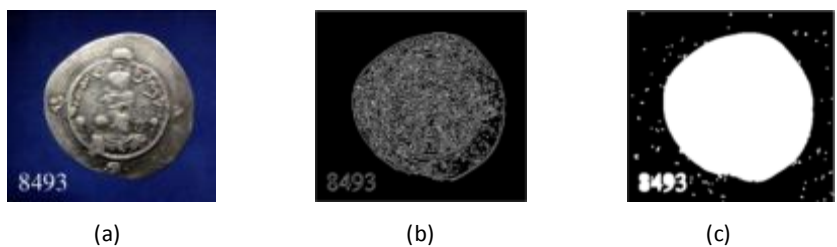

(a)　　　　　　　　　　(b)　　　　　　　　　　(c)

Fig. 2. (a) Coin Image; (b) Edge information with threshold equals to 0.2; (c) Isolated regions

In order to fill these gaps, the Sobel image is dilated using the vertical structuring elements followed by the horizontal structuring elements. At this stage, all holes, which may exist in each object, are filled through the utilization of morphological operators [9]. Fig 2 depicts how mentioned method effectively isolates regions of interest from the background.

Images of ancient coins contain extra information that is coin's identification number. With a view to eliminate this information, which impacts on recognition process, a preprocessing procedure is required. As is evident from Fig. 2 (c), we isolate regions of interest with gradient information in which the end result is a binary image. Now, in order to extract the coin image and eliminate the numbers, the area related to each connected component is calculated and the element with the largest area is kept.

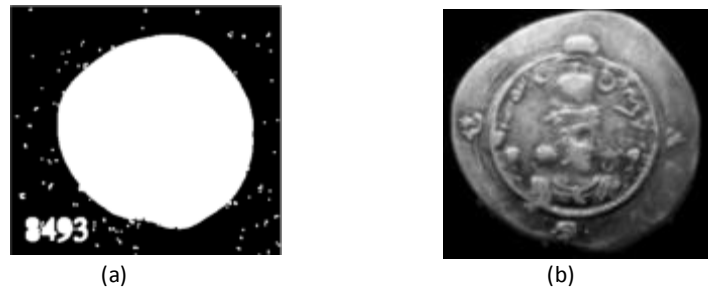

(a)          (b)

Fig. 3. Extracting Region of Interest (ROI); (a) Binary image with different components; (b) extracted coin

Fig 3 shows the end result of preprocessing step, in which the coin is isolated from other parts of the image.

*2.2. EigenCoin*

The aim of this study is to recognize the class of ancient coins in which the Eigen analysis is applied to a set, *S*, of *M* vectorized coin images to construct a manifold of size *N*. Here, *N* is the product of the rows and columns of a normalized coin image, *x*.

$$S = \{x_1, x_2, ..., x_M\} \quad (1)$$

Afterward the mean image ($x$) and the difference between the input image and the mean image (Φ) will be calculated as follows:

$$x = \frac{1}{M} \sum_{n=1}^{M} x_n, \quad \Phi i = x_i - x \quad (2)$$

Subsequent to calculating the mean image, a set of *K* Eigenvectors, which sum of their Eigenvalues covers the maximum energy, are chosen to construct EigenCoin. The $k^{th}$ Eigenvector, $u_k$, is chosen such that:

$$\max \lambda_k; \quad \lambda_k = \frac{1}{M} \sum_{n=1}^{M}(u_k^T \Phi_n)^2$$

s.t.

$$u_l^T u_k = \begin{cases} 1, & l = k \\ 0, & o.w \end{cases} \quad (3)$$

where $u_i$ and $\lambda_i$ are the $i^{th}$ Eigenvector and Eigenvalue, respectively. In recognition step, an input image is projected onto the EigenCoin space and in order to find out its weight with respect to other basis vectors of this manifold, the difference between input image and the mean image gets multiplied by each Eigenvector (4):

$$\Omega^T = [\omega_1, \omega_2, ..., \omega_K],$$
$$\omega_i = u_i^T (x_{input} - x), \quad i = 1, 2, ..., K \quad (4)$$

It is worth pointing out in this context that the fundamental difference between standard Eigenfeature method [10] and proposed one is in calculating Bhattacharyya distance instead of Euclidian distance for the recognition step. It is evident that each vector in EigenCoin space is a multivariate Gaussian distribution and Bhattacharyya distance is calculated as beneath:

$$D_B = \frac{1}{8}(\omega_{input} - \omega_i)^T P^{-1}(\omega_{input} - \omega_i) + \frac{1}{2}\ln\frac{detP}{\sqrt{det\omega_{input} \times det\omega_i}} \quad (5)$$

$$P = \frac{\omega_{input} + \omega_i}{2}, i = 1, 2, \dots, K$$

where $D_B$ is the Bhattacharyya distance, $m_i$, and $P_i$ are the mean and covariance of the distributions, respectively.

### 3. Experimental Results

In order to demonstrate the performance of the proposed method, two sets of experiments are conducted. In the first set, different numbers of Eigenvector with Bhattacharyya distance are examined in order to achieve the maximum recognition rate. In the second set, EigenCoin is compared with Bi-Directional PCA, Wavelet Decomposition, and Harris corner detector which utilizes Bhattacharyya distance as recognition metric. As it is mentioned above, we face the problem of imbalanced data, which causes over-fitting disorder. To solve this and to show the performance of EigenCoin, two measurements are computed which are defined as follows below:

- With the purpose of evaluating the over-fitting problem of PCA, normalized mean-square error (MSE) is computed with the result that the MSE is minimal. Given the first K Eigenvectors, we can obtain the corresponding projector $\omega_K$ and represent the reconstructed vector $\tilde{x}$ as:

$$\tilde{x} = \bar{x} + \omega_K \omega_K^T (x - \bar{x}) \quad (6)$$

The MSE on the training set $MSE_K^{train}$ is defined as:

$$MSE_K^{train} = \frac{\sum_{i=1}^{M}\|x_i^{train} - \tilde{x}_i^{train}\|_2}{\sum_{i=1}^{M}\|x_i^{train} - \bar{x}^{train}\|_2} \quad (7)$$

where $x_i^{train}$ is the $i$th training samples, $\tilde{x}_i^{train}$ is the reconstructed vector of $x_i^{train}$, and $\bar{x}^{train}$ is the mean vector of all training samples.

- To overcome the imbalanced data problem, a weighted metric for calculating the precision of classifiers is defined. In this metric, recognition rate of each class gives multiply by a coefficient and the overall recognition rate is obtained as follows:

$$Precision = \frac{\sum_{i=1}^{4}\alpha_i \times R_i}{\sum_{i=1}^{4}\alpha_i} \propto \frac{1}{C_i} \quad (8)$$

where $\alpha_i$ and $R_i$ are the $i$th coefficient and recognition rate, respectively.

*3.1. Data Set*

We conduct our experiments on Sassanid coins images database. This data set consists of 1288 color images in JPEG format with the resolution of 1772×1553 pixels. These images are distributed in four classes and each class belongs to a Sassanid king. Although each class contains images of both sides of each coin, it should be noted that just obverse images are utilized in the conducted experiments. As is evident in Fig 4, all images subject to preprocessing due to printed demographic data.

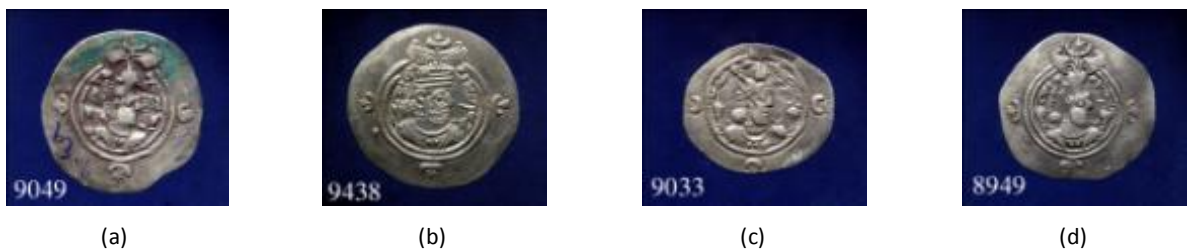

(a)  (b)  (c)  (d)

Fig. 4. Sassanid Coins of (a) Hormozd IV, (b) Hormozd V, (c) Khosrow I, (d) Khosrow II. Coins can be split up into five fundamental elements which are: (1) King's bust, (2) rim(s), (3) three star-moon patterns placed at the sides of coins (4) legends, and (5) a crown pattern at top of king's head

Regulation coefficient and distribution of coins in each class are tabulated in Table 1.

Table 1. Distribution of coins in each class, size of training and testing sets, and $\alpha_i$

| King Name | Training Number | Testing Number | Overall | $\alpha_i$ |
|---|---|---|---|---|
| **Khosrow I** | 35 | 16 | 51 | **3** |
| **Khosrow II** | 343 | 147 | 490 | **1** |
| **Hormozd IV** | 70 | 29 | 99 | **2** |
| **Hormozd V** | 3 | 1 | 4 | **4** |

### 3.2. EigenCoin and Bhattacharyya distance

In this experiment, Eigen analysis with different Eigenvectors is applied to the 451 training vectorized images to construct EigenCoin. Afterwards each testing image is mapped onto new manifold and assigned to a class utilizing Bhattacharyya distance.

With the aim of achieving the best classification accuracy, different number of Eigenvectors is tested and the best result is obtained in a situation in which the number of Eigenvectors is 112. Fig 5 (a) would seem to indicate the overall classification accuracy with respect to various numbers of Eigenvectors. In Fig 5 (b) the mean score error with respect to the different Eigenvectors' number is shown and the confusion matrix is illustrated in Fig 5 (c).

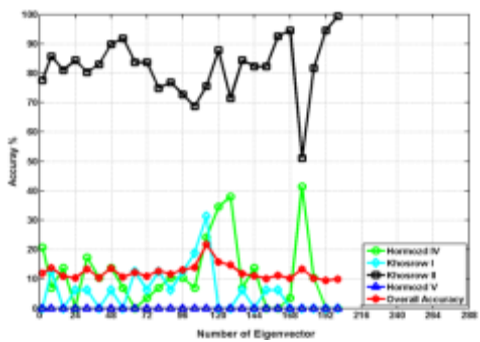
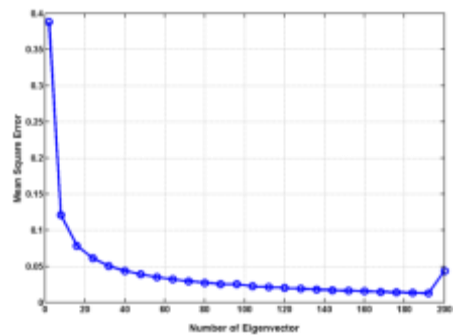

(a)  (b)

|  | Hormozd IV | Khosrow I | Khosrow II | Hormozd V |
|---|---|---|---|---|
| Hormozd IV | 7 | 2 | 20 | 0 |
| Khosrow I | 1 | 5 | 10 | 0 |
| Khosrow II | 14 | 21 | 111 | 1 |
| Hormozd V | 0 | 0 | 1 | 0 |

Confusion Matrix

(c)

Fig. 5. (a) Classification accuracy; (b) MSE; (c) Confusion Matrix

It should be noted that there is a tradeoff between the number of Eigenvectors and computational time. Indeed, the optimal value of Eigenvectors can be obtained on the basis of trials with the result that guarantees both classification accuracy and computational complexity. As is obvious from Fig 5 (a), classification results of images which are related to Khosrow II have significant difference with the result of other classes. The reason is that the distribution of data in the mentioned class is in a way that the constructed manifold lacks the ability to recognized images from other classes. To solve this problem, we utilize the defined precision metric (8) and achieve a reasonable precision. Table 2 shows accuracy of coin classification with respect to different Eigenvectors. In order to save space in this table, just eight highest precisions are tabulated.

Table 2. Accuracy of coin classification with respect to different Eigenvectors. $C_i$ is an acronym for the name of a class

| Eigenvector<br>Accuracy (%) | 8 | 32 | 48 | 104 | 112 | 120 | 128 | 176 |
|---|---|---|---|---|---|---|---|---|
| $C_1$ | 6.9 | 17.2 | 13.8 | 6.9 | **24.1** | 34.5 | 37.9 | 41.4 |
| $C_2$ | 12.5 | 6.25 | 6.25 | 18.7 | **31.2** | 0 | 0 | 0 |
| $C_3$ | 85.7 | 80.3 | 89.8 | 68.7 | **75.5** | 87.7 | 71.4 | 5 |
| $C_4$ | 0 | 0 | 0 | 0 | **0** | 0 | 0 | 0 |
| **Overall** | 13.7 | 13.3 | 13.6 | 13.9 | **21.7** | 15.8 | 14.7 | 13.4 |

*3.3. Competitive methods*

In order to evaluate the proposed method and demonstrate its efficiency, accuracy of EigenCoin is compared with three state of the art feature extraction and object classification methods. These competitive methods are Bi-Directional PCA (BDPCA) with Assembled Matrix Distance (AMD) [6], Wavelet decomposition [7], and Harris corner detector [8] with Bhattacharyya distance.

*3.3.1. Bi-Directional PCA and Assembled Matrix Distance*

There is a fundamental difference between PCA and BDPCA in analyzing and extracting the feature of interest. BDPCA preserves the structure of data, i.e., the input data should not be vectorized. In this method, two transformations so-called row and column scatter matrices are utilized to extract directly the feature matrix. In recognition step, a kind of matrix norm called Assembled Matrix Distance is performed on the testing data to classify it properly.

In order to evaluate the BDPCA in the coin classification problem, six pairs of row and column scatter matrices are investigated. The best result is obtained through the utilization of 15 rows and 35 columns which is 19.54%. Refer to Fig 6 for more details.

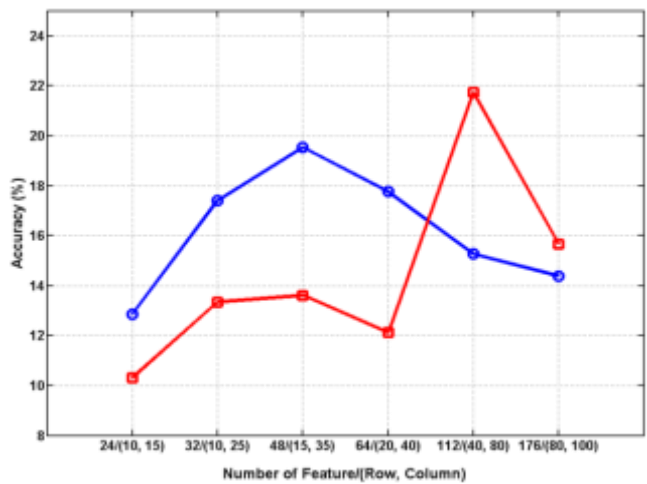

Fig. 6. Accuracy of BDPCA with different number of features in comparison with EigenCoin

As is obvious in Figure 6, when the number of rows and columns of the scatter matrix increase, the overall accuracy decreases. In contrast, the proposed method provides reasonably high recognition rate also on those occasions in which the number of feature vectors increases.

*3.3.2. Wavelet and Bhattacharyya Distance*

Wavelet decomposition is a non-linear transformation with the advantage over other non-linear ones for which both frequency and location information are captured simultaneously. This transformation is worked by means of a low pass and band pass filter. The low pass filter constructs the approximate image and the band pass filter constructs details images. In the conducted experiment, four levels of decomposition with Haar filter are observed along which feature vectors with 5, 17, 65, and 257 entries are obtained. The first and second entries of each feature vector relates to the mean and standard deviation of approximate image while the other entries are the standard deviation of details images.

After constructing feature vector, the Bhattacharyya distance is utilized to calculate the distance between test image and each training ones. Finally, the entry with minimum distance leads the recognition step to find the proper class of a test image. The accuracy of coin classification with this method is tabulated in Table 3.

Table 3. Accuracy of Wavelet and EigenCoin methods with different numbers of features

| Number of Features | 5 | 8 | 16 | 17 | 64 | 65 | 112 | 257 |
|---|---|---|---|---|---|---|---|---|
| Wavelet accuracy (%) | 17.11 | - | - | 17.46 | - | 18.08 | - | **19.17** |
| EigenCoin accuracy (%) | - | 13.7 | 13.89 | - | 12.11 | - | **21.75** | - |

As is clear in this table, even the feature vector with size of 257 has not the capability of EigenCoin in identifying the class of input images.

*3.3.3. Harris Corner Detector and Bhattacharyya Distance*

It is reported by Arandjelovic [2] that local feature descriptors such as Harris [8], SUSAN [11], Hough transform [12], SIFT [13], SURF [14], and GLOH [15] have the capability of classifying ancient coins better than other investigated methods in modern coins. Harris corner detector considers the differential of the corner score with respect to its direction. This operator applies Eigen analysis to a matrix which expresses the gradient distribution of a point's neighbors and computes the magnitude of Eigenvalues. Then, a corner will be detected in the event that there are two large positive Eigenvalues at a point. Results of applying Harris corner detector are shown in Fig. 7.

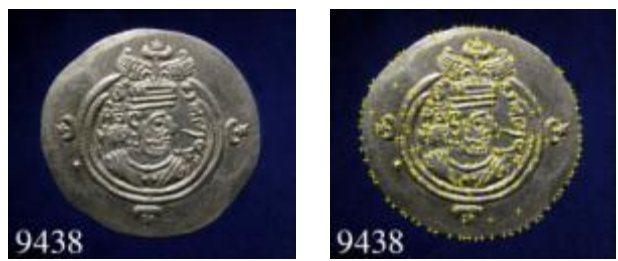

Fig. 7. Results of applying Harris corner detector

In the conducted experiment, the feature vector consists of corner point's intensity. It was observed in the experiments that the feature vector has the dimension between 103 and 184. After extracting the feature vector, Bhattacharyya distance between a new image and those that construct the feature space is calculated for using in classification task. The detection percent related to each class and overall accuracy are tabulated in Table 4. As is evident, the overall accuracy is 18.81% which is not comparable with the best result of EigenCoin method, nonetheless the classification results of Hormozd IV is fairly better than the proposed method.

Table 4. Classification result of applying Harris corner detector. Here, D is an abbreviation for dimensional.

| Method \ Class | Hormozd IV | Khosrow I | Khosrow II | Hormozd V | Overall |
|---|---|---|---|---|---|
| **Harris (184D)** | 37.93% | 12.5% | 74.83% | 0% | **18.82%** |
| **EigenCoin (112D)** | 24.1% | 31.2% | 75.5% | 0% | **21.75%** |

## 4. Conclusion

In recent years, much research has been devoted to the automation of pattern recognition applications; however, few have harnessed the cultural heritage applications. In fact, it seems that the causal factor relates to the problems of existing databases, i.e., small sample size, imbalanced data, etc. In this paper, we explored *EigenCoin*, new manifold for Sassanid coin classification. Continuing with this rationale, our methodology for classifying ancient coins is obviously promising in imbalanced data set in which we conducted our experiments on it.

In EigenCoin, first, we choose 70% of data and apply Eigen analysis to the isolated coins in order to construct a manifold. Afterwards, a number of Eigenvectors are chosen based on their Eigenvalue to construct EigenCoin space. Along these same lines, we map each test image onto EigenCoin and calculate Bhattacharyya distance between the test image and all train images. Finally, based on the obtained distances, the test coin will be assigned to one of four predefined classes, i.e., Khosrow I, Khosrow II, Hormozd IV, and Hormozd V. In order to demonstrate the accuracy and efficiency of the proposed method, we tested not only different numbers of Eigenvectors and calculated different normalized mean square errors, but also investigated the effect of Bi-Directional PCA, Wavelet decomposition, and Harris Corner detector.

It was observed in the course of experiments that the EigenCoin outperformed other examined methods and achieved the accuracy from 9.45% up to 21.75%. Furthermore, results of calculating normalized mean-square error prove that EigenCoin has the capability of handling the over-fitting problem in the imbalanced Sassanid coin database. In our opinion, it is not an unjustifiable assumption that the classification rate of each class is reported separately. Therefore, a metric is defined to balance the overall accuracy percentage.